\documentclass[10pt,twocolumn,letterpaper]{article}

\usepackage[pagenumbers]{wacv} 

\usepackage{graphicx}
\usepackage{amsmath}
\usepackage{amssymb}
\usepackage{booktabs}
\usepackage{multirow}
\usepackage{xcolor}
\usepackage[accsupp]{axessibility} 

\usepackage[pagebackref,breaklinks,colorlinks]{hyperref}
\usepackage{verbatim}

\usepackage[capitalize]{cleveref}
\crefname{section}{Sec.}{Secs.}
\Crefname{section}{Section}{Sections}
\Crefname{table}{Table}{Tables}
\crefname{table}{Tab.}{Tabs.}

\def\wacvPaperID{2526} 
\def\confName{WACV}
\def\confYear{2024}

\begin{document}


\title{Complex Organ Mask Guided Radiology Report Generation}
\author{Tiancheng Gu, Dongnan Liu, Zhiyuan Li, Weidong Cai\\
University of Sydney, Sydney, Australia\\
{\tt\small $\begin{Bmatrix}tigu8498,zhli0736\end{Bmatrix}$@uni.sydney.edu.au, 
$\begin{Bmatrix}dongnan.liu,tom.cai\end{Bmatrix}$@sydney.edu.au}}

\maketitle

\begin{abstract}
    The goal of automatic report generation is to generate a clinically accurate and coherent phrase from a single given X-ray image, which could alleviate the workload of traditional radiology reporting.However, in a real-world scenario, radiologists frequently face the challenge of producing extensive reports derived from numerous medical images, thereby medical report generation from multi-image perspective is needed.In this paper, we propose the \textbf{C}omplex \textbf{O}rgan \textbf{M}ask \textbf{G}uided~(termed as COMG) report generation model, which incorporates masks from multiple organs~(e.g., bones, lungs, heart, and mediastinum), to provide more detailed information and guide the model's attention to these crucial body regions. Specifically, we leverage prior knowledge of the disease corresponding to each organ in the fusion process to enhance the disease identification phase during the report generation process. Additionally, cosine similarity loss is introduced as target function to ensure the convergence of cross-modal consistency and facilitate model optimization.Experimental results on two public datasets show that COMG achieves a $11.4\%$ and $9.7\%$ improvement in terms of BLEU@4 scores over the SOTA model KiUT on IU-Xray and MIMIC, respectively. The code is publicly available at \url{https://github.com/GaryGuTC/COMG_model}.
\end{abstract}

\section{Introduction}
\label{sec:introduction}

\begin{figure}[ht]
\begin{center}
\includegraphics[width=1.0\linewidth]{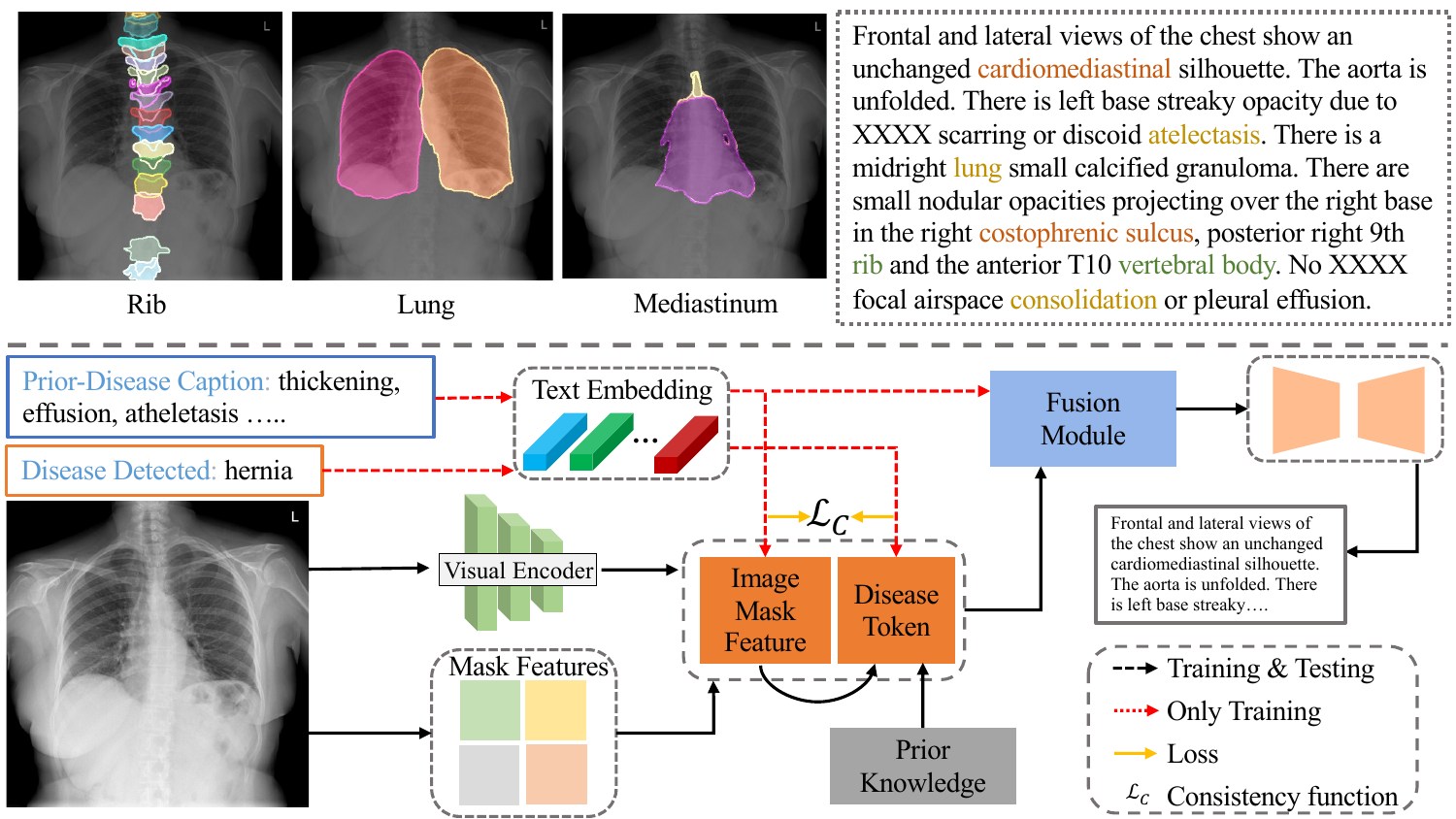}
\end{center}
\vspace{-0.3cm}
\caption{Top: the above section display the relationship between the mask images and the captions, that includes findings related to bone (green), lung (yellow), and mediastinum (red). Bottom: the below section illustrates the basic structure of the COMG model. Best viewed by zooming in.}
\label{fig:introduction}
\end{figure}


Radiology image analysis plays a pivotal role in disease detection \cite{radiology_importance}. In clinical practice, it is time-consuming, costly, and error-prone for radiologists to review numerous radiology images and generate the corresponding reports for further analysis. To mitigate this challenge, there has been a growing interest in exploring automated radiology report generation (RRG) techniques \cite{show_tell, AdaAtt, att2in}. Recently, deep learning methods have been widely explored to generate a textual analysis for given images (e.g., image captioning) \cite{flamingo_model, meshed_memory, show_attn_tell}. However, RRG differs from the standard image caption task and is more complex and challenging \cite{xpronet}. Radiology images focus on some specific regions related to the disease, which account for only a small fraction of the entire image \cite{dcl}. The main difference between radiographs is the specific area associated with the disease. Such challenges also remain in the text descriptions for RRG, i.e., the differences are mainly from the analysis of the diseases, while the descriptions for the normal tissues are similar.


Several RRG techniques have been proposed to solve the challenges from various perspectives and achieve appealing performance. Specifically,~\cite{R2Gen, R2GenCMN, r2gencmm_rl, xpronet} have been developed by improving the encoder-decoder structure and enhancing the feature fusion across different modalities. By incorporating prior knowledge from auxiliary resources, such as the region detection prediction, and retrieval of textual information,~\cite{repsnet, rgrg_model, dcl} are proposed to further enhance the report generation via the comprehensive knowledge. Despite their outstanding performance, none of these methods consider the pixel-level information for the specific tissues. As indicated in~\cite{ChestXRayAnatomySegmentation}, the pixel-level segmentation masks for the tissues are also critical for further analysis in radiology images, which reflects the semantic correlations of the tissue, as indicated in Fig.~\ref{fig:introduction}. For example, the masks can indicate the size and shape of the tissue, as well as their spatial distributions, which are related to the recognition and understanding of diseases~\cite{dodia2022recent}. Since such critical information has not been considered for RRG, existing methods are suboptimal by missing such pixel-level semantic tissue correlations for radiology images.

In this work, we propose a \textbf{C}omplex \textbf{O}rgan \textbf{M}ask \textbf{G}uided (COMG) framework for radiology report generation which considers the dense pixel-level information of X-ray images. Specifically, pre-trained segmentation models (e.g., CXAS \cite{ChestXRayAnatomySegmentation}) are employed to extract segmentation masks for key tissues correlated to the disease diagnosis in the generated reports (e.g., heart, bones, etc.), as indicated in Fig.~\ref{fig:introduction}. The mask information provides the model with local-level semantic information on the organs/tissues during report generation, including the morphological structures of specific tissues, and the spatial relationships among different tissues. Then, the feature prototypes are extracted according to the masks for different organs. In addition, we further incorporate the disease keywords associated with each specific tissue as text-level guidance, whose feature embeddings are fused with the feature prototypes to further enhance the understanding of the correlations between each specific tissue and the corresponding disease during the report generation learning. Finally, cross-modal consistency mechanisms are developed to facilitate feature extraction at the vision and language levels by inducing their similarities.

Our contributions can be summarized as follows: 1) We propose to improve the report generation by enhancing the understanding of the semantic correlations of the tissues in the radiology images via mask information; 2) We introduce the disease keywords associated with each tissue as the text-level prior knowledge to further enhance the tissue feature learning; 3) To facilitate the feature extractions for the image and text, two cross-modal consistency mechanisms are developed; 4) Our proposed COMG method is indicated to be effective by achieving outstanding performance on two public medical report generation benchmarks.

\section{Related Works}
\subsection{Image Caption}
Image caption tasks aim to generate text descriptions based on the input images. Early works~\cite{anderson2018bottom, AdaAtt, xlinear, object_detection} are developed based on the Long Short-Term Memory (LSTM) model \cite{lstm} and the Convolutional Neural Network (CNN) model \cite{cnn}. Recently, the transformer models based on the attention module \cite{transformer} have been widely employed due to their outstanding ability to process the vision and language features \cite{vision_transformer, meshed_memory, xlinear, metransformer}. Despite the appealing performance of these methods in general image caption benchmarks, their applicability is limited for radiology report generation, which is challenging due to the data bias between the normal and disease tissues at the image and text levels~\cite{metransformer}.


\subsection{Radiology Report Generation}

In recent years, several deep learning-based methods have been proposed for radiology report generation, which can be mainly categorized into two types: 1) improving the model structure \cite{Expert_defined_Keywords, CDGPT2, R2Gen, R2GenCMN, r2gencmm_rl}, and 2) incorporating external modality information \cite{repsnet, rgrg_model, dcl}.

CDGPT2 \cite{CDGPT2} proposes to fuse visual and semantic features by concatenating them into a multi-modal decoder. AlignTransformer \cite{aligntransformer} proposed to use the multi-grained transformer (MGT) to improve multi-modal features fusion. In addition, R2Gen \cite{R2Gen} and R2GenCMN \cite{R2GenCMN} models are developed to enhance the model structure for filtering and fusing image and caption information separately using the LSTM\cite{lstm} and CMN modules. R2GenCMM-RL \cite{r2gencmm_rl} further improves the R2GenCMN via reinforcement learning. XproNet \cite{xpronet} model proposed to initialize the models via prototype matrix initialization, and a multi-label contrastive loss function to guide the optimization process.

Additionally, some works have started to explore leveraging external information to facilitate the report generation process. The RepsNet model \cite{repsnet} utilizes external information from a pre-trained VQA model on the VQA-radiology dataset \cite{vqa_rad}. It proposes to integrate the features for the answer information from the VQA model with image features for accurate report generation. The RGRG model \cite{rgrg_model} is developed to first utilize the bounding boxes to detect abnormal regions in the image. Then, some suitable detected areas are chosen for report generation, instead of using the entire image. The DCL model \cite{dcl} integrates information from a pre-constructed knowledge graph that contains correlations between caption words. Such information is further processed by a dynamic graph encoder and then combined with the image features using a blip-like structure \cite{blip} to generate accurate reports via the comprehensive understanding of the disease words. In this work, we propose to incorporate a new type of external knowledge, which is the segmentation masks for the key issues related to the disease mentioned in the text analysis. To the best knowledge, we are making an early attempt to improve the RRG tasks via the pixel-level semantic knowledge for the tissues in the radiology images.

\begin{figure*}[ht]
\begin{center}
\includegraphics[width=0.9\linewidth]{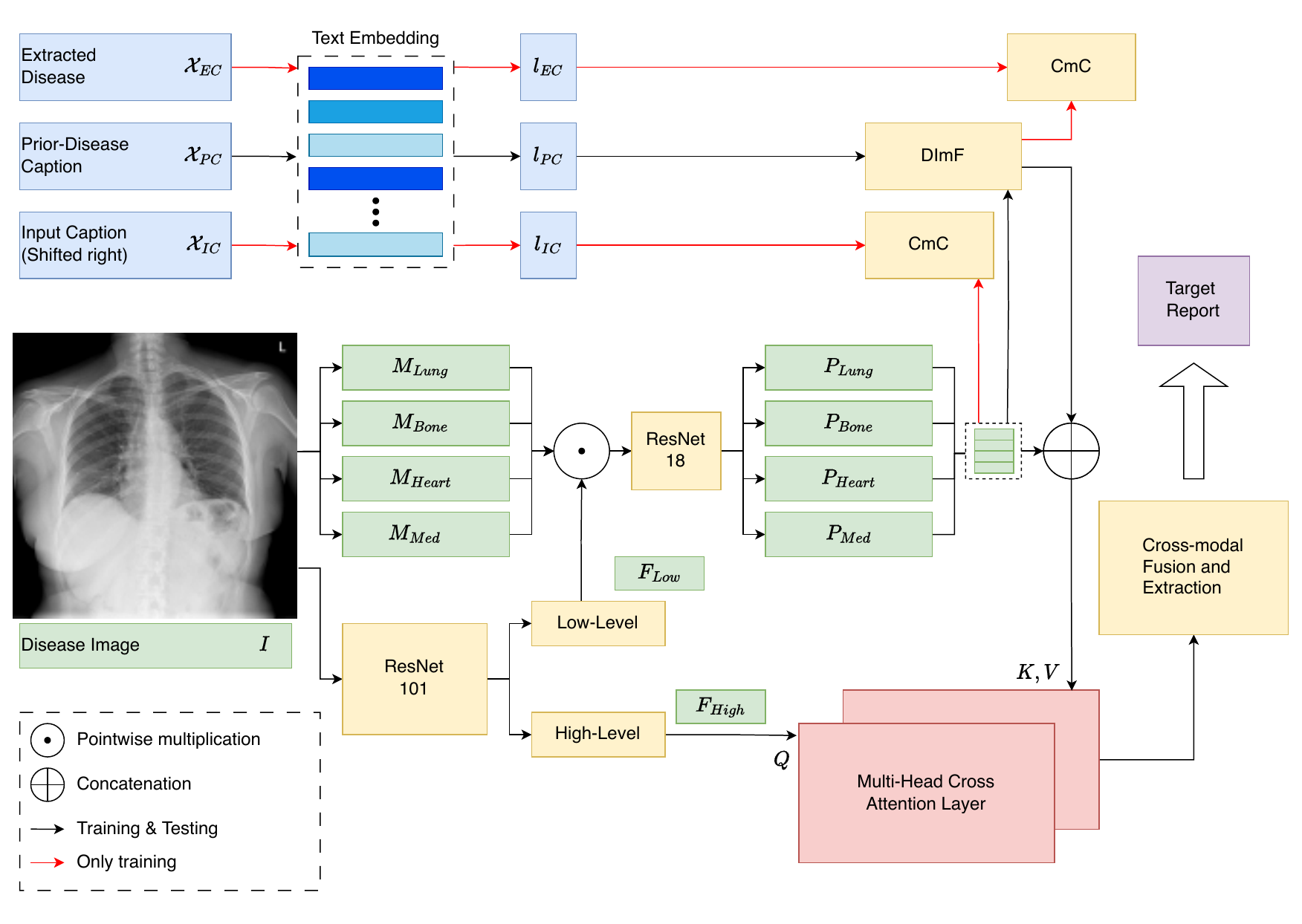}
\end{center}
\vspace{-0.5cm}
\caption{The overall architecture of our proposed COMG model. The \textit{DImF} represents the fusion mechanisms between the embeddings for the prior-disease captions and the mask-guided prototype features, with details shown in Fig.~\ref{fig: the structure of image mask fusion and disease image mask fusion}, and \textit{CmC} is the cross-modal consistency. Finally, the encoder-decoder structure generates the report. More detailed information is provided in Sec.~\ref{sec:Propsoed Method}.}
\label{fig: The Strucutre of the COMG model}
\end{figure*}
\section{COMG}
\label{sec:Propsoed Method}
The overview of our proposed \textbf{C}omplex \textbf{O}rgan \textbf{M}ask \textbf{G}uided (COMG) method is shown in Fig. \ref{fig: The Strucutre of the COMG model}. Our method is established on the R2GenCMN model \cite{R2GenCMN}, which is constructed by an encoder-decoder structure using the transformer model \cite{transformer} and multi-modal feature fusion mechanism. Our main contributions include the Mask-guided Organ Prototype Feature Extraction mechanism (Sec.~\ref{sec:Mask-guided_Organ}), the Cross-modal Correlation Studies between the Tissues and the Diseases (Sec.~\ref{sec:image part}), and the Multi-modal Feature Fusion and Consistency Mechanisms (Sec.~\ref{sec:caption part}).

\subsection{Mask-guided Organ Prototype Feature Extraction}
\label{sec:Mask-guided_Organ}

Due to most diseases existing on the body organs, the pixel-level mask information provided will help the model to particularly recognize these key areas \cite{tissue_segmentation}. To this end, we use the pre-trained CXAS model \cite{ChestXRayAnatomySegmentation}, which was trained on the PAX-RAY++ segmentation dataset \cite{paxray_dataset}. By inferring the model on each X-ray image, it generates segmentation masks for various organs, including the heart, ribs, lung, and mediastinum (more specific information about masks is shown in the Supplementary Material). However, it's noteworthy that the COMG model is evaluated on two public benchmarks (IU-Xray, MIMIC-CXR), and the diseases mentioned in the reports are mainly related to four organs: bone, lung, heart, and mediastinum. Therefore, only partial masks belonging to these categories for each image will be employed. Since some organs are only related to limited types of disease, extracting the pixel-level masks for each organ and employing them for further analysis separately can induce the model to learn the correct correlations between the organs and disease, while ignoring negative pairing relationships.

After obtaining the masks for the key organs, we propose to extract the prototype features for each organ. The overall process is indicated in Fig.~\ref{fig: The Strucutre of the COMG model} and:
\begin{equation}
P_{og} = R_{ref} (F_{Low} \odot M_{\textit{og}}),
\label{equation:prototype}
\end{equation}
where $og \in \{Bone, Lung, Heart, Mediastinum\}$. Specifically, the input images first pass through a ResNet 101 feature extractor for the intermediate features $F_{Low}$, and the final features $F_{High}$. Next, the resized $F_{Low}$ is multiplied pointwise by the masks pre-extracted from each organ category ($M_{\textit{og}}$). These multiplied features further pass through a ResNet18 ($R_{ref}$) for refinement to obtain the prototype features for the four key tissues. Based on the segmentation masks, these features contain semantic information for different key organs related to report generation (i.e., organ shapes and spatial relationships).


\subsection{Cross-modal Correlation Studies between the Tissues and the Diseases}
\label{sec:image part}

\begin{figure}[ht]
\begin{center}
\includegraphics[width=1\linewidth]{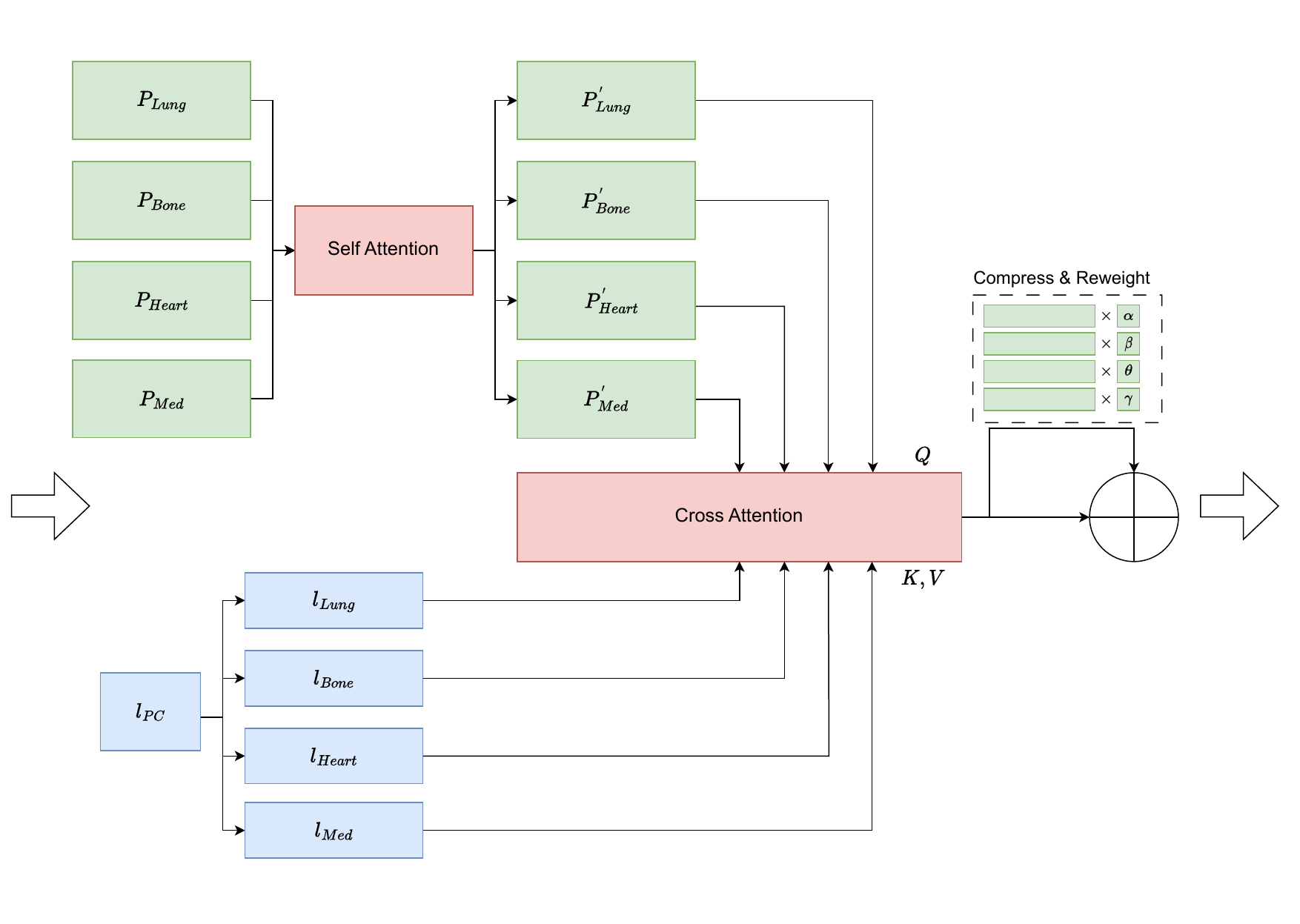}
\end{center}
\vspace{-0.5cm}
\caption{The structure of the disease image-mask fusion (\textit{DImF}) module. The prototype features ($P'$) of each organ will be inputted into a decoder structure with the corresponding disease token space ($\l$) separately. \textbf{Reweight} are four learnable parameters to be multiplied with each corresponding organ's features. In addition, $\bigoplus$ means concatenation and \textbf{Compress} is the feature dimension compression. More specific information is provided in Sec.~\ref{sec:image part}.}
\label{fig: the structure of image mask fusion and disease image mask fusion}
\end{figure}

To accurately distinguish between diseases and normal cases, we combine the prototype features of each organ with the features from the keywords for the related diseases. The process is illustrated in Fig. \ref{fig: the structure of image mask fusion and disease image mask fusion}. Referring to the disease symptom graph \cite{KiUT} (more details in the Supplementary Material), we can obtain the corresponding prior disease captions related to each organ. Note that these captions are based on the predefined knowledge graph and do not related to the report annotations of the images. Fig.~\ref{fig: The Strucutre of the COMG model} and Fig. \ref{fig: the structure of image mask fusion and disease image mask fusion}, the correlation token for each organ ($Tok_{og}$) is calculated via:

\begin{equation}
Tok_{og} = CA\big(SA(P_{og}), l_{og}, l_{og}\big).
\label{equation:dimf}
\end{equation}
Specifically, the prototype features for each organ ($P_{og}$ in Eq.~\ref{equation:prototype}) first pass through a self-attention $SA$ layer. Next, the prior captions for each disease pass through transformer encoders for disease keyword embeddings. Within each organ type, the processed prototype features and the prior-disease caption embeddings $l_{og}$ are fused via cross-attention mechanisms $CA(Q, K, V)$ to obtain the cross-modal correlation tokens $Tok_{og}$ for the organ and its corresponding diseases. Specifically, the processed class-wise prototype features $SA(P_{og})$ are employed as the query, while the corresponding prior-disease caption embeddings $l_{og}$ as the key and values. In addition, four (one for each tissue class) learnable parameters are developed to re-weight the importance of each cross-modal correlation token $Tok_{og}$. The tokens $Tok_{og}$ from four organs are fused together according to these learnable parameters for a global correlation token $Tok_{glb}$.

\subsection{Multi-modal Feature Fusion and Consistency}
\label{sec:caption part}

\paragraph{Multi-modal Feature Fusion.} To facilitate the report generation learning via the comprehensive information aforementioned, we propose to fuse the global-level features directly extracted from the input images with the multi-modal features from the $DImF$ module. As indicated in Fig.~\ref{fig: The Strucutre of the COMG model}, the output of the $DImF$ module $Tok_{glb}$ is firstly reshaped and concated with each Organ Prototype Feature $P_{og}$. These concated features are employed as the key and value for the multi-head cross-attention layer, while the global-level image features $F_{high}$ are the query. Such fused features contain knowledge from multiple perspectives, including the dense mask for key tissues, the text descriptions for the diseases, and the global-level features for the whole images. By passing them jointly with the embeddings for the extracted disease keywords from the report to transformer-based decoders, the quality of the generated report can be further improved.

\paragraph{Cross-modal Consistency.} To further facilitate the feature extraction process under multiple modalities, we propose cross-modal consistency mechanisms ($CmC$ module in Fig.~\ref{fig: The Strucutre of the COMG model}). First, we propose to maximize the similarity between the embeddings of the input captions $l_{IC}$ and the mask-guided organ prototype feature $P_{og}$ (obtained from Sec.~\ref{sec:Mask-guided_Organ}). According to the analysis of existing RRG methods~\cite{R2Gen,rgrg_model}, the analysis in the reports is correlated to the specific regions and tissues in the radiology images. To this end, under the ideal feature extraction scenario, the organ prototype features should be dependent on the report descriptions. In addition, we also induce the similarity learning between the feature embeddings for the extracted disease for each image ($l_{EC}$), and the cross-modal correlation tokens ($Tok_{og}$ from Sec.~\ref{sec:image part}). For the cross-modal correlation tokens containing the text-level information regarding the diseases based on prior knowledge and the image-level information based on masks, their similarity with the disease keywords extracted from the ground truth should be maximized. It is because, under the optimal situation, the features for the multimodal tokens and the text keywords are both about the high-level characteristics of the diseases. For each similarity learning process, the two features are firstly resized into the same scale, then the cosine similarity loss ($L_{sim}$)~\cite{consine_similarity} is utilized to enlarge their similarities. Specifically, the $L_{sim}$ is defined as $L_{sim}(a, b)  = 1 - \frac{a^T b}{||a||~||b||}$.


\begin{table*}[!t]
\centering
\resizebox{\linewidth}{!}{
\begin{tabular}{l|c|c|cccccc}
\hline
Datase & Methods            & YEAR & BLEU@1 & BLEU@2 & BLEU@3 & BLEU@4 & METEOR & ROUGEL \\
\hline
\multirow{12}{*}{IU-Xray}   & R2Gen\cite{R2Gen}  &  2020& 0.470   & 0.304   & 0.219   & 0.165   & 0.187  & 0.371  \\
                            & SEBTSAT+KG \cite{SENTSAT_KG} & 2020& 0.441 & 0.291 & 0.203 & 0.147 & - & 0.304  \\
                            & PPKED\cite{ppked}    & 2021& 0.483   & 0.315   & 0.224   & 0.168   & 0.190  & 0.376   \\
                            & CMCL\cite{CMCL}    &  2022& 0.473   & 0.305   & 0.217   & 0.162   & 0.186  & 0.378  \\
                            & JPG\cite{jpg}      &  2022& 0.479   & 0.319   & 0.222   & 0.174   & 0.193  & 0.377  \\
                            & CMM+RL\cite{r2gencmm_rl} & 2022& 0.475   & 0.309   & 0.222   & 0.170   & 0.191  & 0.375  \\
                            & KiUT\cite{KiUT}    &  2023& \underline{0.525}   & \underline{0.360}   & \underline{0.251}   & \underline{0.185}   & \textbf{0.242}  & \textbf{0.409}  \\
                            & METransformer\cite{metransformer} & 2023  & 0.483   & 0.322   & 0.228   & 0.172   & 0.192  & 0.380  \\
                            & DCL\cite{dcl}       &  2023& -        & -        & -        & 0.163   & 0.193  & 0.383\\
                            & R2GenCMN$^{*}\dag$ \cite{R2GenCMN} & 2022  & 0.470   & 0.304   & 0.222   & 0.170   & 0.191  & 0.358  \\
                            \cline{2-9}
                            & \textbf{COMG}  & \textbf{Ours}   & 0.482   & 0.316   & 0.233   & 0.184   & 0.198  & 0.382  \\
                            & \textbf{COMG + RL(Ours)} & \textbf{Ours}  & \textbf{0.536}   & \textbf{0.378}   & \textbf{0.275}   & \textbf{0.206}   & \underline{0.218}  & \underline{0.383}  \\
                            \hline
\multirow{13}{*}{MIMIC-CXR} & M2Transformer \cite{m2transformer} & 2020 & 0.332   & 0.210   & 0.142   & 0.101   & 0.134  & 0.264 \\
                            & R2Gen\cite{R2Gen}  & 2020& 0.353   & 0.218   & 0.145   & 0.103   & 0.142  & 0.277 \\
                            & PPKED\cite{ppked}  & 2021& 0.360   & 0.224   & 0.149   & 0.106   & 0.149  & 0.284 \\
                            & CMCL\cite{CMCL}    & 2022& 0.344   & 0.217   & 0.140   & 0.097   & 0.133  & 0.281 \\
                            & CMM+RL\cite{r2gencmm_rl} & 2022  & 0.353   & 0.218   & 0.148   & 0.106   & 0.142  & 0.278  \\
                            & UAR\cite{uar}  & 2023 & 0.363   & 0.229   & 0.158   & 0.107   & \underline{0.157}  & \underline{0.289}  \\
                            & KiUT\cite{KiUT} & 2023  & \textbf{0.393}   & \textbf{0.243}   & \underline{0.159}   & \underline{0.113}   & \textbf{0.160}  & 0.285  \\
                            & DCL\cite{dcl}  & 2023   & -       & -       & -       & 0.109   & 0.150  & 0.284  \\
                            & R2GenCMN$^{*}\dag$\cite{R2GenCMN} & 2022& 0.348   & 0.206   & 0.135   & 0.094   & 0.136  & 0.266  \\
                            \cline{2-9}
                            & \textbf{COMG}     & \textbf{Ours} & 0.346    & 0.216   & 0.145   & 0.104   & 0.137   & 0.279 \\
                            & \textbf{COMG + RL(Ours)} & \textbf{Ours} & \underline{0.363}   & \underline{0.235}   & \textbf{0.167}   & \textbf{0.124}   & 0.128   & \textbf{0.290}  \\
                            \hline
\end{tabular}
}
\caption{The results of the COMG model and other tested models in IU-Xray (upper part) and MIMIC-CXR (lower part) datasets. $^{*}$ indicates that we tested the results ourselves, which may differ from the results reported in the original papers of other models. $\dag$ denotes the baseline model. The results for other models were obtained from their original papers. The best result is presented in bold, and the second-best result is underlined.}
\label{table:compare_with_other_model}
\end{table*}

\subsection{Training and Inference Details}

The COMG model is optimized in two stages. The overall loss function $\mathcal{L}_{\textit{T1}}$ for the first stage is three folds, defined as:
\begin{equation}
	\mathcal{L}_{\textit{T1}} =  \mathcal{L}_{\textit{CE}} + \beta \mathcal{L}_{Sim_{\textit{IM}}} + \theta \mathcal{L}_{Sim_{\textit{DT}}},
	\label{equation:loss function}
\end{equation}
where \textit{CE} is the cross-entropy loss for report generation study following~\cite{R2Gen}, $Sim_{\textit{IM}}$ means the similarity-maximization loss between the embeddings of the input captions and the mask-guided organ prototype feature, and the $Sim_{\textit{DT}}$ is the cosine similarity loss between the cross-modal correlation tokens and extracted disease keywords features. We add tradeoff parameters to these loss functions to balance the overall optimization process, with $\beta$ and $\theta$ set as $0.1$. We have also presented the experimental analysis regarding different selections in the following sections.

After optimized via $\mathcal{L}_{\textit{T}}$ for the first stage, we propose another stage of optimization for better performance by incorporating reinforcement learning (RL). Specifically, we included an additional BLEU score as a reward for RL to improve sentence coherence, combined with $\mathcal{L}_{\textit{CE}}$ for report generation.

During inference, the cross-modal correlation tokens are first extracted from each image. Note that it is accessible since these tokens can be acquired by incorporating the image features with the masks from the pre-train segmentation models and the prior-disease captions from the pre-defined knowledge graph, which do not require ground-truth annotations. Then, the tokens are integrated with the image-level features via the multi-head attention. Finally, the decoder receives such fused features as input for report predictions.

\section{Experiments}
In this section, we first introduce the details of the experimental settings, including datasets, baseline models, and evaluation metrics. We then conduct the proposed COMG model on two datasets and evaluated it alongside some state-of-the-art approaches. In addition, ablation studies and the hyperparameters analysis of the COMG model are further presented.

\subsection{Experiment Settings}

\subsubsection{Datasets}
\label{sec:dataset}

Two widely studied RRG benchmarks are employed to test the COMG model: IU-Xray \cite{IU-xray} from Indiana University and MIMIC-CXR \cite{mimic_cxr} from the Beth Israel Deaconess Medical Center. The MIMIC-CXR dataset is the largest publicly available radiography dataset, with 473,057 chest X-ray images and 206,563 associated reports. The IU-Xray is a relatively smaller dataset, which contains 7,470 chest X-ray images and 3,955 corresponding reports. Both datasets are divided into training, testing, and validation sets in a ratio of 7:2:1. More details of the datasets can be referred to the Supplementary Material. For both datasets, we followed Chen et al. \cite{R2GenCMN} to pre-process captions and images. Before entering the model, each original radiology image was resized to 3 * 224 * 224 and normalized. In comparison, the mask images were resized to 1 * 224 * 224 to fit the mid-process fusion with mid-image features. To increase the model's robustness, images and masks were also enhanced with random cropping and random horizontal flipping. The captions were cleaned up, including removing punctuation and converting some words that appear less than three times to the token $<unk>$.

\subsubsection{Baseline and Evaluation Metrics}
\label{sec:baseline}


\paragraph{Baseline.} We compare the COMG model with nine existing radiology report generation models that have state-of-the-art (SOTA) results in the IU-Xray dataset. These models include R2Gen \cite{R2Gen}, PPKED \cite{ppked}, CMCL \cite{CMCL}, KIUT \cite{KiUT}, and METransformer \cite{metransformer}. We also compare the COMG model with the R2GenCMN \cite{R2GenCMN} baseline model (marked $\dag$ in Table \ref{table:compare_with_other_model}).

In addition to the IU-Xray dataset, we compare the COMG model with nine SOTA models on a different dataset MIMIC-CXR, including M2Transformer \cite{m2transformer}, UAR \cite{uar}, DCL \cite{dcl}, and the baseline model ``R2GenCMN''. The results of other comparison methods are cited from their respective papers, while the results of R2GenCMN are re-implemented by us as the baseline model (marked as $^{*}$ in Table \ref{table:compare_with_other_model}).


\paragraph{Evaluation Metrics.} We evaluate the quality of our report generation using natural language generation (NLG) metrics, \ie BLEU~[1-4] \cite{bleu}, METEOR \cite{meteor} and ROUGE-L \cite{rouge}. These metrics measure the similarity between the generated caption and the ground truth in terms of word-level n-grams. In addition, we follow the approach of \cite{R2Gen, R2GenCMN, r2gencmm_rl} and use clinical efficacy (CE) metrics to evaluate the reports generated on the MIMIC-CXR dataset with their corresponding target captions. The CE metrics assess the presence of a set of significant clinical observations that can capture the diagnostic accuracy of the generated reports.

\subsubsection{Implementation Details}
\label{sec:implementation details}
We choose the ResNet101 pre-trained on ImageNet as the image extractor model and the CXAS~\cite{ChestXRayAnatomySegmentation} pre-trained on the radiology segmentation dataset PAX-RAY++~\cite{paxray_dataset} as the mask extractor model. Our model is trained on a single NVIDIA GeForce RTX 3090 GPU with a 24GB memory.
For optimization, following~\cite{R2GenCMN}, we use the Adam optimizer~\cite{Adam}. The initial learning rates for the ResNet101 feature extractor and other components are set to 1e-4 and 5e-4, separately. During the inference stage, we incorporate the beam search~\cite{beam_search} into the COMG model, with a step setting of 3. More experiment information has been provided in the Supplementary Material.

\begin{table}[!t]
\begin{tabular}{l|ccc}
\hline
\multirow{2}{*}{Method} & \multicolumn{3}{c}{CE Metric} \\
\cline{2-4} & \multicolumn{1}{c}{Precision} & \multicolumn{1}{c}{Recall} & \multicolumn{1}{c}{F1} \\
\hline
R2Gen \cite{R2Gen}              & 0.333  & 0.273 & 0.276 \\
CMM+RL \cite{r2gencmm_rl}       & 0.342  & 0.294 & 0.292 \\
METransformer \cite{metransformer}   & 0.364 & \underline{0.309}  & 0.311 \\
KiUT \cite{KiUT}   & \underline{0.371}   & \textbf{0.318}  & \underline{0.321}  \\
R2GenCMN$^{*}\dag$ \cite{R2GenCMN}       & 0.334  & 0.275 & 0.278 \\
\hline
\textbf{COMG(Ours)}    & \textbf{0.424}  &  0.291   &  \textbf{0.345}       \\
\hline
\end{tabular}
\caption{Comparison of clinical efficacy metrics for the MIMIC-CXR dataset. These metrics measure the accuracy of clinical abnormality descriptions. The best result is presented in bold, and the second-best result is underlined.}
\label{table:comp_CE_Metric}
\end{table}
\subsection{Experiment Results and Analysis}
\subsubsection{Radiology Report Generation}
\label{sec:compare_with_other_model}

Two evaluation metrics are used for comparison: conventional natural language generation (NLG) metrics and clinical efficacy (CE) metrics. These are common metrics used to evaluate the report generation task. The results are shown in Table \ref{table:compare_with_other_model} and Table \ref{table:comp_CE_Metric}, respectively.

\begin{table*}
\scriptsize
\resizebox{\linewidth}{!}{
\begin{tabular}{l|l|ccccccc|ccccccc}
\hline
\multirow{2}{*}{\#} & \multirow{2}{*}{Method} & \multicolumn{7}{c|}{IU-Xray}  & \multicolumn{7}{c}{MIMIC-CXR}  \\
\cline{3-9} \cline{10-16}
                & & B@1 & B@2 & B@3 & B@4 & MET. & RGL. & AVG. & B@1 & B@2 & B@3 & B@4 & MET. & RGL. & AVG.\\
                \hline
\multirow{2}{*}{1} & \multirow{2}{*}{Baseline} & \multirow{2}{*}{0.462} & \multirow{2}{*}{0.299} & \multirow{2}{*}{0.221} & \multirow{2}{*}{0.172} & \multirow{2}{*}{0.193} & \multirow{2}{*}{0.37}  & \multirow{2}{*}{-} & \multirow{2}{*}{0.330} & \multirow{2}{*}{0.201} & \multirow{2}{*}{0.134} & \multirow{2}{*}{0.094} & \multirow{2}{*}{0.134} & \multirow{2}{*}{0.269} & \multirow{2}{*}{-} \\
\multicolumn{1}{l|}{} & \multicolumn{1}{l|}{} & \multicolumn{6}{l}{} & \multicolumn{1}{l|}{} & \multicolumn{7}{l}{}\\
\hline
\multirow{2}{*}{2} & \multirow{2}{*}{+ Mk}  & \multirow{2}{*}{0.504} & \multirow{2}{*}{0.323} & \multirow{2}{*}{0.227} & \multirow{2}{*}{0.170} & \multirow{2}{*}{0.192} & \multirow{2}{*}{0.388} & \multirow{2}{*}{$\downarrow 1.1\%$} & \multirow{2}{*}{0.338} & \multirow{2}{*}{0.206} & \multirow{2}{*}{0.138} & \multirow{2}{*}{0.098} & \multirow{2}{*}{0.130} & \multirow{2}{*}{0.270} & \multirow{2}{*}{$\uparrow 4.3\%$} \\
\multicolumn{1}{l|}{} & \multicolumn{1}{l|}{} & \multicolumn{6}{l}{} & \multicolumn{1}{l|}{} & \multicolumn{7}{l}{}\\
\hline
\multirow{2}{*}{3} & \multirow{2}{*}{+ Mk + Sim$_{\textit{IM}}$} & \multirow{2}{*}{0.469} & \multirow{2}{*}{0.303} & \multirow{2}{*}{0.223} & \multirow{2}{*}{0.175} & \multirow{2}{*}{0.187} & \multirow{2}{*}{0.367}  & \multirow{2}{*}{$\uparrow 1.7\%$} & \multirow{2}{*}{0.343} & \multirow{2}{*}{0.211} & \multirow{2}{*}{0.141} & \multirow{2}{*}{0.100} & \multirow{2}{*}{0.134} & \multirow{2}{*}{0.275} & \multirow{2}{*}{$\uparrow 6.4\%$} \\
\multicolumn{1}{l|}{} & \multicolumn{1}{l|}{} & \multicolumn{6}{l}{} & \multicolumn{1}{l|}{} & \multicolumn{7}{l}{}\\
\hline
\multirow{2}{*}{4} & + Mk + DT & \multirow{2}{*}{0.484} & \multirow{2}{*}{0.320} & \multirow{2}{*}{0.234} & \multirow{2}{*}{0.182} & \multirow{2}{*}{0.204} & \multirow{2}{*}{0.379}  & \multirow{2}{*}{$\uparrow 5.8\%$} & \multirow{2}{*}{0.347} & \multirow{2}{*}{0.213} & \multirow{2}{*}{0.142} & \multirow{2}{*}{0.102} & \multirow{2}{*}{0.136} & \multirow{2}{*}{0.275} & \multirow{2}{*}{$\uparrow 8.5\%$}\\
\multicolumn{1}{l|}{} & + Sim$_{\textit{DT}}$ & \multicolumn{6}{l}{} & \multicolumn{1}{l|}{} & \multicolumn{7}{l}{}\\
\hline
\multirow{2}{*}{5} & + Mk + Sim$_{\textit{IM}}$& \multirow{2}{*}{0.482} & \multirow{2}{*}{0.316} & \multirow{2}{*}{0.233} & \multirow{2}{*}{0.184} & \multirow{2}{*}{0.198} & \multirow{2}{*}{0.382} & \multirow{2}{*}{$\uparrow 7.0\%$} & \multirow{2}{*}{0.347} & \multirow{2}{*}{0.216} & \multirow{2}{*}{0.145} & \multirow{2}{*}{0.104} & \multirow{2}{*}{0.137} & \multirow{2}{*}{0.279} & \multirow{2}{*}{$\uparrow 10.6\%$}  \\
\multicolumn{1}{l|}{} & + DT + Sim$_{\textit{DT}}$ & \multicolumn{6}{l}{} & \multicolumn{1}{l|}{} & \multicolumn{7}{l}{}\\
\hline
\multirow{2}{*}{6} & + Mk + Sim$_{\textit{IM}}$ & \multirow{2}{*}{0.536} & \multirow{2}{*}{0.378} & \multirow{2}{*}{0.275} & \multirow{2}{*}{0.206} & \multirow{2}{*}{0.218} & \multirow{2}{*}{0.383} & \multirow{2}{*}{$\uparrow 20.0\%$} & \multirow{2}{*}{0.363} & \multirow{2}{*}{0.235} & \multirow{2}{*}{0.167} & \multirow{2}{*}{0.124} & \multirow{2}{*}{0.128} & \multirow{2}{*}{0.290} & \multirow{2}{*}{$\uparrow 31.9\%$}\\
\multicolumn{1}{l|}{} & + DT + Sim$_{\textit{DT}}$ + RL & \multicolumn{6}{l}{} & \multicolumn{1}{l|}{} & \multicolumn{7}{l}{}\\
\hline
\end{tabular}
}
\caption{The ablation study of the COMG model on the IU-Xray and MIMIC-CXR datasets. AVG indicates the improvement in the BLEU@4 value compared to the baseline model, while RL stands for reinforcement learning. MET. and RGL. represent Meteor and Rouge-L, respectively.}
\label{table:ablation_study}
\end{table*}

\paragraph{Descriptive Accuracy.} We report the descriptive accuracy in Table \ref{table:compare_with_other_model}. As can be seen from the results on IU-Xray, the COMG model outperforms the baseline model ``R2GenCMN'' in all aspects, including BLEU~[1,2,3,4], Meteor, and Rouge-L by adding all the contributions mentioned in Sec. \ref{sec:introduction}. By adding reinforcement learning as a second step in training for the COMG model, our model excels in BLEU~[1,2,3,4] and achieves the second best results in METEOR and ROUGE-L. In radiology report generation, BLEU@4 is an important guideline \cite{R2GenCMN}, and the COMG model achieves a significant improvement in this metric compared to ``KiUT'' (\ie, $0.185 \rightarrow 0.206$).

In the MIMIC-CXR dataset, the ``RGRG'' and ``METransformer'' are excluded. The ``RGRG'' used a very large model with a 24-layer decoder, making it difficult for others to reproduce its results. The ``METransformer'' did not make its code public, which makes it impossible to re-run the experiment on the MIMIC-CXR dataset. Furthermore, the dataset split used in our evaluation was different from these two models. Compared to the baseline model ``R2GenCMN'', our model has significant improvements in all six evaluation metrics. By adding RL to the COMG model, our model achieves substantial performance gains in BLEU~[3,4] and ROUGE-L, and achieves the second-best result in BLEU~[1,2] compared to other existing models, \eg, the BLEU@4 score increased by $9.7\%$ compared to the second-best result from ``KiUT''. However, the result of METEOR has dropped slightly when RL is added, because the reward in RL is based on BLEU metrics while ignoring the METEOR score. We have also tried to employ other metrics as rewards for RL, and we noticed their performance is inferior to the RL with BLEU.

\begin{figure*}[ht!]
\begin{center}
\includegraphics[width=.97\linewidth]{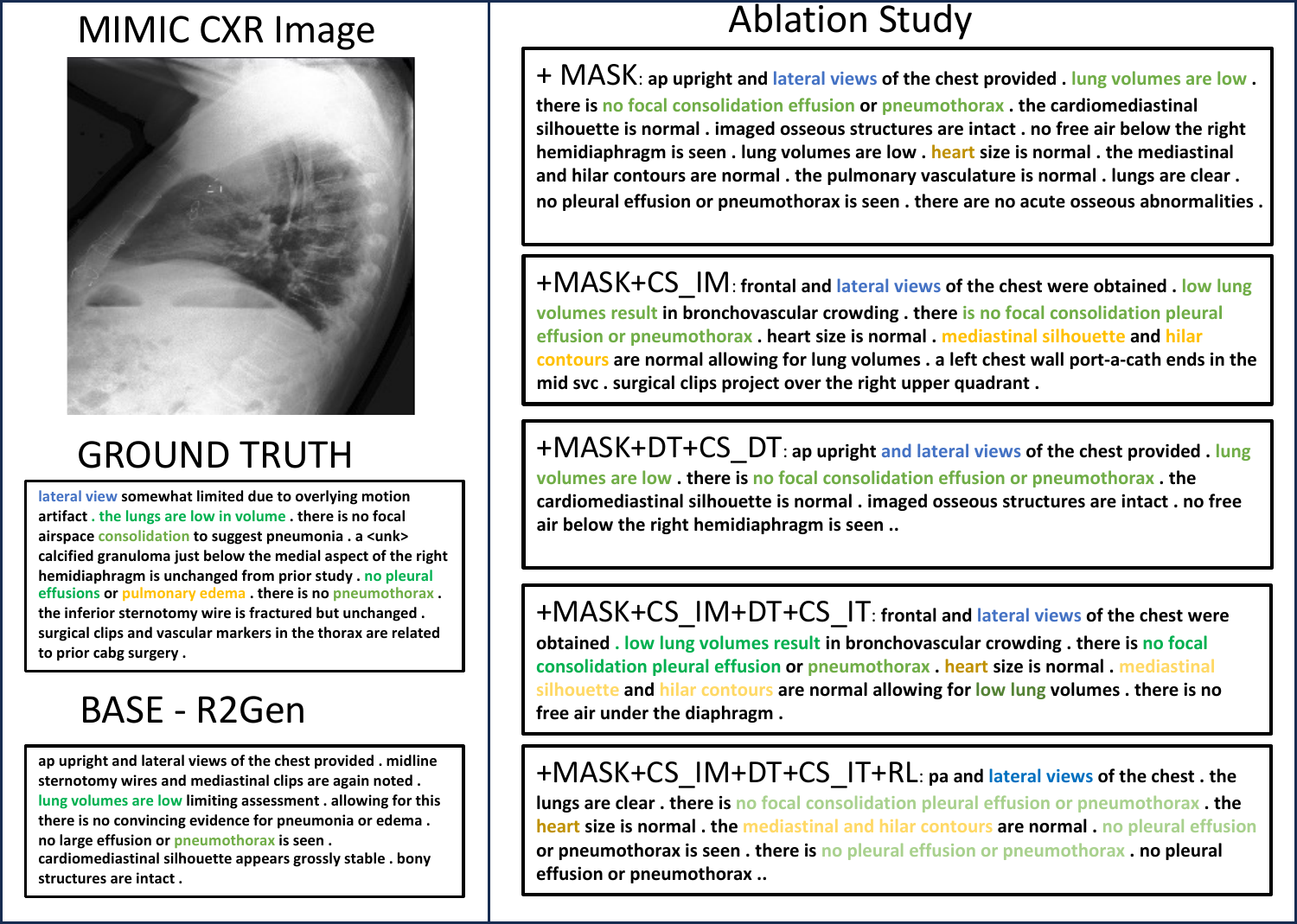}
\end{center}
\vspace{-0.3cm}
\caption{An example of the report generated by different models in the ablation study. The left side of the image displays the input image from the MIMIC-CXR dataset, the corresponding ground truth, and the report generated by the baseline model. In the image, we have marked the keywords of organs and diseases in different colors other than black. More specific information is shown in Sec.~\ref{sec:qualitative_analysis}.}
\label{fig: qualitative_analysis}
\end{figure*}

\paragraph{Clinical Correctness.} Table~\ref{table:comp_CE_Metric} reports the quantitative results of our proposed model and 4 SOTAs, \ie, R2Gen \cite{R2Gen}, MEtransformer \cite{metransformer}, KiUT \cite{KiUT}, and the baseline model R2GenCMN \cite{R2GenCMN}, on the MIMIC-CXR dataset. 
As can be seen, our COMG model performs better in precision and F1 score than all SOTAs, \eg, precision increased by $14.3\%$ and F1 increased by $7.5\%$. Compared to ``R2GenCMN'', COMG achieves a significant improvement on all metrics. 

\subsubsection{Ablation Study}
\label{sec:ablation_study}
In this section, we conduct an ablation study to investigate the effect of each designed module in our approach. Table \ref{table:ablation_study} shows the experimental results on two datasets: IU-Xray and MIMIC-CXR. Specifically, \textit{$MK$} represents the introduction of the mask-guided organ prototype features for model training (Sec.~\ref{sec:Mask-guided_Organ}). \textit{$Sim_{IM}$} represents including the similarity loss $\mathcal{L}_{Sim_{\textit{IM}}}$, while $Sim_{DT}$ represents adding the similarity loss $\mathcal{L}_{Sim_{\textit{DT}}}$. $DT$ represents adding the cross-modal correlation token (Sec.~\ref{sec:image part}), and $RL$ represents the addition of reinforcement learning. We evaluated the model using three kinds of metrics: BLEU~[1,2,3,4], METEOR, and ROUGE-L. The AVG. column shows the average increase of each model compared to the baseline model, based on the BLEU@4 metric, which is the most important evaluation metric in radiology report generation tasks. 


By comparing \#1 and \#2 in Table \ref{table:ablation_study}, it indicates that incorporating the mask-guided organ prototype features has made the greatest contribution to the improvement of the model's performance, which indicates the importance of the pixel-level information for report generation. Additionally, we have also validated the effectiveness of the two cross-modal consistency mechanisms. The performance gain for \#3 over \#2 shows the benefits of enlarging the cross-modal similarity between the organ prototype features and the report descriptions. By jointly employing the cross-modal correlation tokens with its similarity learning with the disease keywords from ground truth, the baseline has been improved by a large margin (\#4). Next, the improvement of the first-stage model \#5 over the baseline \#1 has further indicated the model's effectiveness by jointly integrating all proposed modules. Finally, \#6 shows the result of combining all contributions and adding reinforcement learning. It significantly improves performance in most metrics, while only losing fewer marks of Rouge-L in IU-Xray and Meteor in MIMIC-CXR due to the rewards being set on the BLEU metrics.

\subsubsection{Qualitative Analysis}
\label{sec:qualitative_analysis}

To further investigate the effectiveness of our method, we perform qualitative analysis on the MIMIC-CXR dataset (shown in Fig.~\ref{fig: qualitative_analysis}). In the example, we have highlighted the keywords related to organs and diseases in distinct colors for clear differentiation. It shows that if the model detects certain parts of the disease incorrectly, its prediction will fail to generate the corresponding descriptions successfully. More specifically, baseline can only detect lung and pneumothorax, while our COMG can detect more details about mediastinal and hilar contours, which makes our report more vivid and accurate. More examples can be found in the Supplementary Material.

\subsubsection{Hyper-parameter Analysis}
\label{sec:Hyper-parameter Analysis}

\begin{table}
\resizebox{\linewidth}{!}{
\begin{tabular}{c|ccccccc}
\hline
\multicolumn{1}{c|}{} & \multicolumn{7}{c}{Loss Coefficient} \\ 
\hline
$\beta$ & 0.1 & 0.1 & 1 & 1 & 10 & 1 & 10\\
\hline
$\theta$ & 0.1 & 1 & 0.1 & 1 & 1 & 10 & 10\\
\hline
B@1  & \textbf{0.482} & 0.444 & 0.433 & 0.470 & 0.468 & 0.435 & 0.418\\
B@2  & \textbf{0.316} & 0.282 & 0.273 & 0.306 & 0.300 & 0.178 & 0.274\\
B@3  & \textbf{0.233} & 0.200 & 0.190 & 0.213 & 0.215 & 0.198 & 0.197\\
B@4  & \textbf{0.184} & 0.148 & 0.142 & 0.158 & 0.162 & 0.150 & 0.148\\
MET. & \textbf{0.198} & 0.174 & 0.177 & 0.197 & 0.190 & 0.178 & 0.174\\
RGL. & \textbf{0.382} & 0.341 & 0.342 & 0.366 & 0.365 & 0.360 & 0.360\\ 
\hline
\end{tabular}
}
\caption{The influence of each coefficient on each loss component in the loss function. This experiment was conducted using the IU-Xray dataset.}
\label{table:analysis_coefficient}
\end{table}

Table~\ref{table:analysis_coefficient} shows the results of different combinations of loss function coefficients $\beta$, and $\theta$. We changed the coefficients of $\beta$ and $\theta$ from 0.1 to 10 to evaluate the influence of each loss function coefficient. Table \ref{table:analysis_coefficient} lists the results of our COMG under this range. We find that 0.1, and 0.1 are the best choices for $\beta$, and $\theta$, respectively. 


\section{Conclusion}
In this paper, we propose a novel COMG method for generating precise radiology reports. It employs complex organ masks to provide pixel-wise semantic information for accurate report generation. Additionally, it incorporates disease keywords linked to each tissue, utilizing them as text-level prior knowledge to further refine tissue feature learning. To streamline the feature extraction process for both images and text, we have developed two cross-modal consistency mechanisms to enhance feature learning accuracy. Our method has been tested on two popular benchmarks, and the results show its effectiveness in generating accurate and meaningful reports. In future works, we plan to enhance the COMG’s ability to recognize abnormal tissues/regions by incorporating external resources (e.g., Chest ImaGenome dataset \cite{ImaGenome_Dataset}). This can further improve the reports’ quality on disease recognition and understanding.


\section{Supplementary Material}
\subsection{Methodology}
\label{Appendix:methodology}

\subsubsection{Details on Tissue Segmentation Masks}
\label{Appendix:mask_information}

\begin{figure}[ht]
\begin{center}
\includegraphics[width=1\linewidth,height=0.3\textwidth]{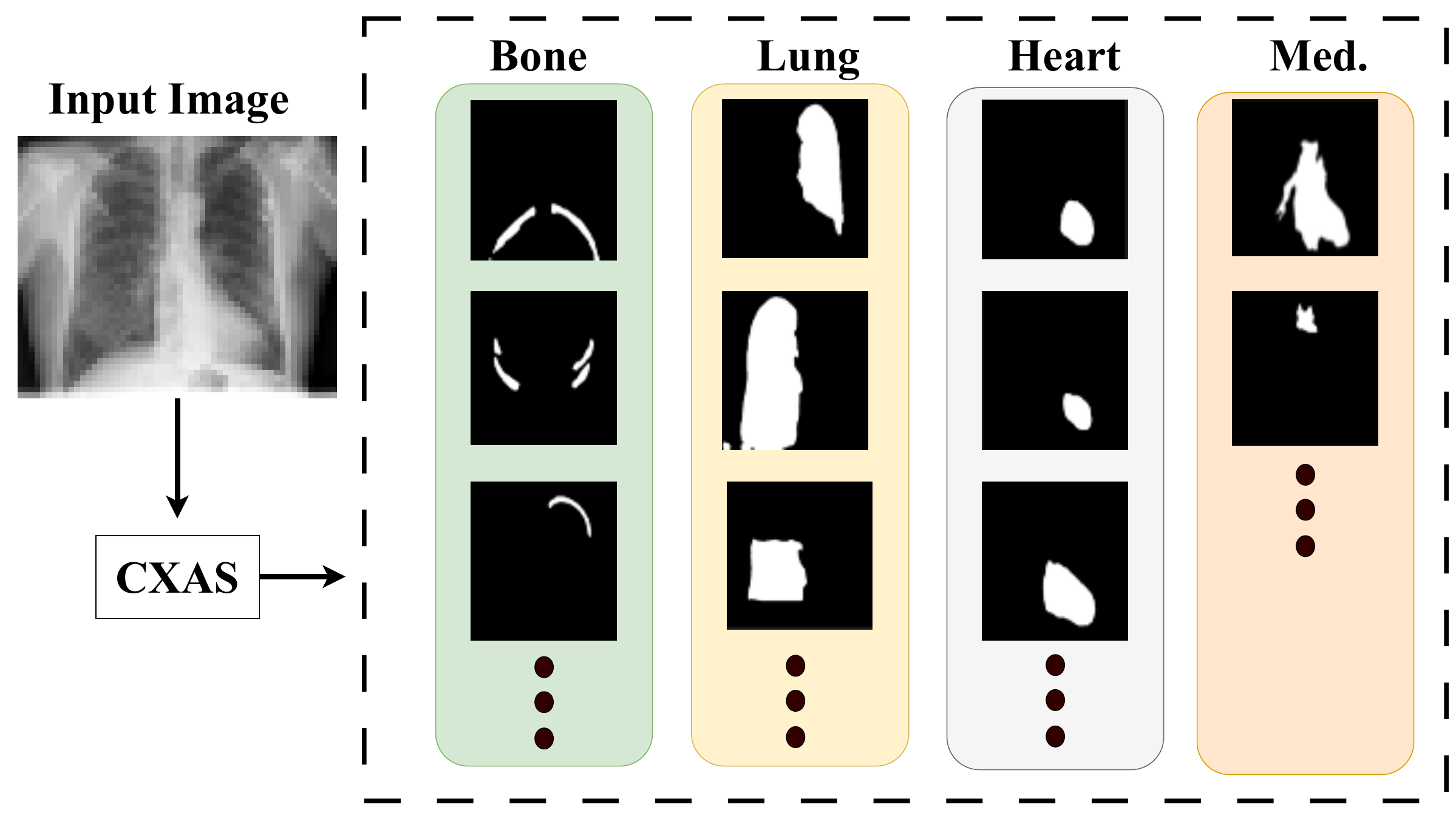}
\end{center}
\caption{The mask image generated after the CXAS from the MIMIC-CXR benchmark. Med. means the mediastinum.}
\label{appendix:fig_Image_mask}
\end{figure}

We provide more details regarding the tissue segmentation masks extracted by the pre-train CXAS model \cite{ChestXRayAnatomySegmentation} for the proposed Mask-guided Organ Prototype Feature Extraction mechanism (Sec.~\ref{sec:Mask-guided_Organ}). The details are presented in Table \ref{Appendix:table_mask_info}. For each given image, the CXAS model can extract totally a fixed number of masks for different tissues in the radiology images, such as lung lobes, lung zones, and lung halves. However, some tissues are irrelevant to the descriptions in the report, such as the abdomen, trachea, etc. Therefore, out of a total of 159 generated mask images, only 97 are useful for the COMG model. These include 70 for bones, 15 for lungs, 6 for the heart, and 6 for the mediastinum. Fig. \ref{appendix:fig_Image_mask} shows the example of mask images extracted from the input image using the CXAS model.

\subsubsection{Disease Symptom Graph}
\label{Appendix:disease_info}

\begin{figure}[ht]
\begin{center}
\includegraphics[width=0.9\linewidth,height=0.25\textwidth]{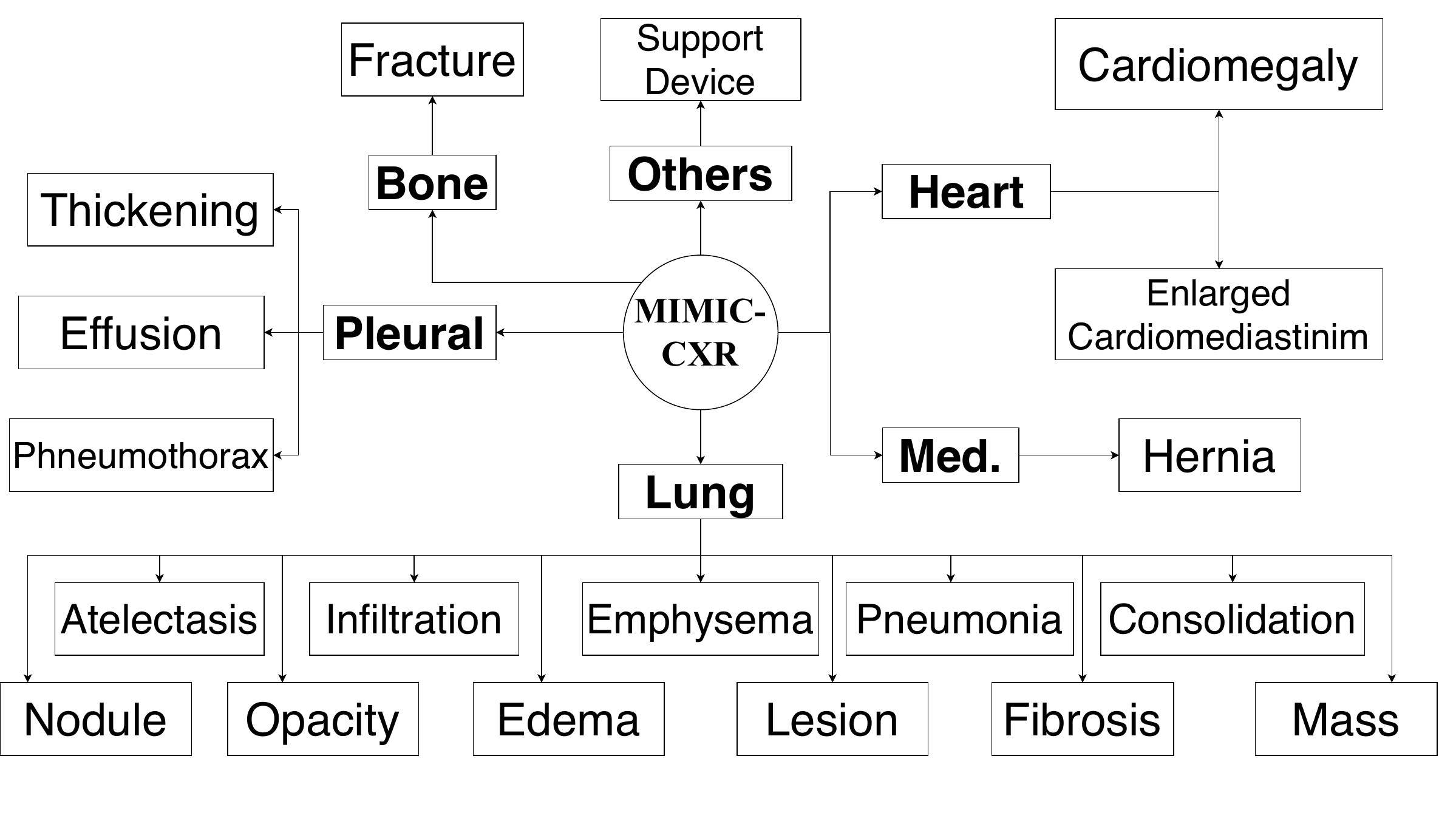}
\end{center}
   \caption{The symptom graph summarizes the related diseases for each organ in the MIMIC-CXR dataset. Specifically, we focused on four organs: bone, lung, heart, and mediastinum.}
\label{appendix_fig:disease_tokens}
\end{figure}

We have presented the details on the disease knowledge graph obtaining the prior-disease captions, as presented in Fig. \ref{appendix_fig:disease_tokens}. It is based on the professional analysis of the relationship between the organs and the corresponding diseases in the radiology images. Specifically, the graph was developed in \cite{KiUT}, taking into account symptom correlation, symptom characteristics, occurrence location, and other relevant factors.

\begin{table}
\begin{tabular}{llcclll}
\hline
\multicolumn{1}{l|}{Regional Mask Images} & \multicolumn{1}{l|}{Num.} & \multicolumn{1}{l|}{Region} & \multicolumn{1}{l}{Total Mask} \\ 
\hline
\multicolumn{1}{l|}{Lung lobes}    & \multicolumn{1}{l|}{5}  & \multicolumn{1}{c|}{\multirow{3}{*}{Lung}} & \multirow{7}{*}{159} \\
\multicolumn{1}{l|}{Lung zones}    & \multicolumn{1}{l|}{8}  & \multicolumn{1}{c|}{}                      &      \\
\multicolumn{1}{l|}{Lung halves}   & \multicolumn{1}{l|}{2}  & \multicolumn{1}{c|}{}                      &      \\ \cline{1-3}
\multicolumn{1}{l|}{Heart region}  & \multicolumn{1}{l|}{6}  & \multicolumn{1}{c|}{Heart}                 &      \\ \cline{1-3}
\multicolumn{1}{l|}{Mediastinum}   & \multicolumn{1}{l|}{6}  & \multicolumn{1}{c|}{Mediastinum}   &    \\ \cline{1-3}
\multicolumn{1}{l|}{Ribs}          & \multicolumn{1}{l|}{46} & \multicolumn{1}{c|}{\multirow{2}{*}{Bone}} &    \\
\multicolumn{1}{l|}{Ribs super}    & \multicolumn{1}{l|}{24} & \multicolumn{1}{c|}{}                      &           \\ \cline{1-3}
\multicolumn{1}{l|}{...}           & \multicolumn{1}{l|}{...}& \multicolumn{1}{l|}{...}                   & \multicolumn{1}{l}{} \\
\hline
\end{tabular}
\caption{The specific information of masks generated by the CXAS model \cite{ChestXRayAnatomySegmentation}, as well as the mask images we ultimately used.}
\label{Appendix:table_mask_info}
\end{table}

\begin{figure*}[ht]
\begin{center}
\includegraphics[width=1\linewidth]{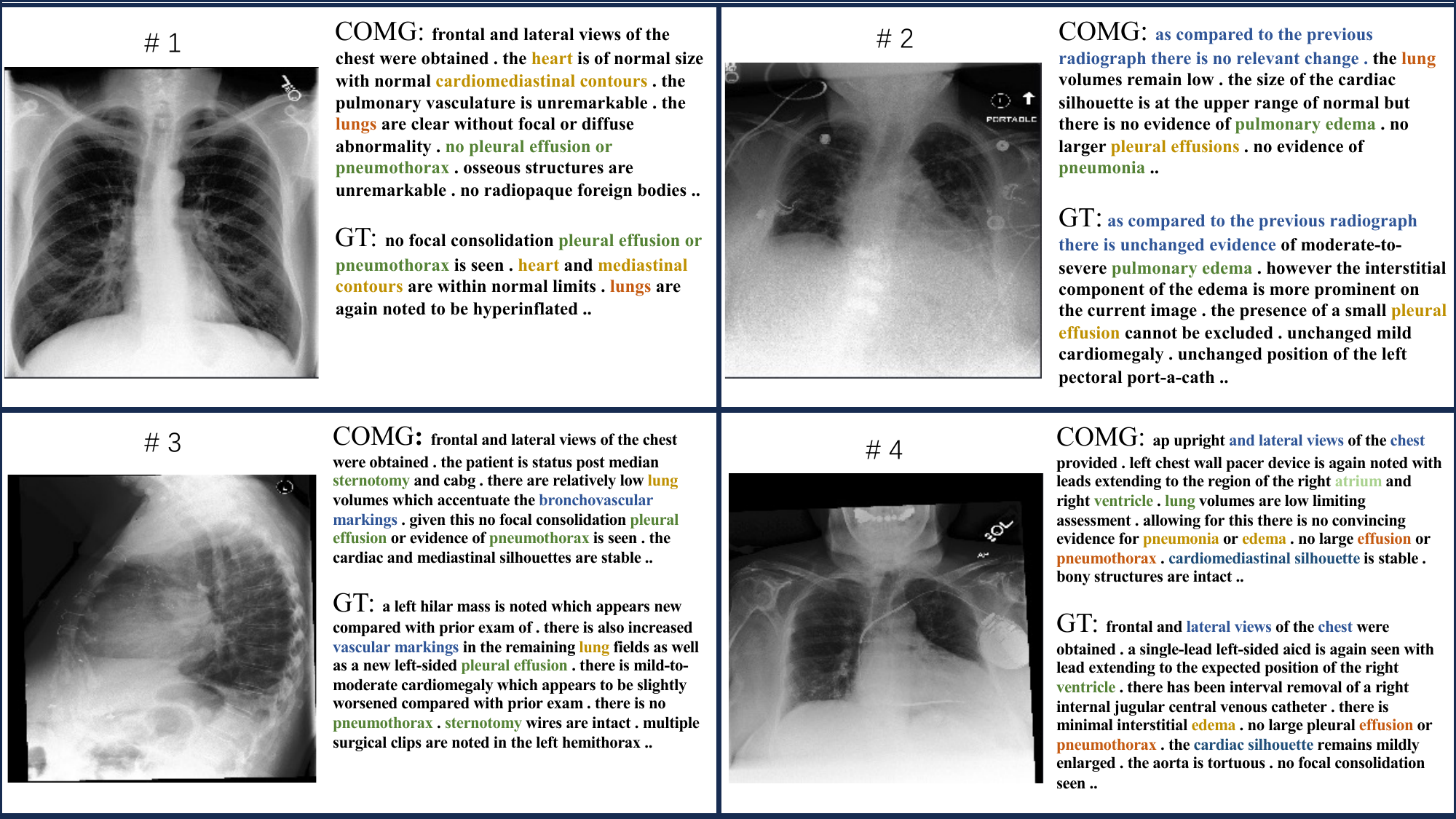}
\end{center}
\vspace{-0.3cm}
\caption{The visulization of prediction results by the COMG model. GT is the abbreviation of the Ground Truth.}
\label{appendix_fig:visualization_prediction}
\end{figure*}

\subsection{Dataset}
\label{Appendix:Dataset_info}

Table \ref{table:dataset} provides detailed information on the two benchmarks used to evaluate the COMG model. The table shows that MIMIC-CXR contains more cases than IU-Xray. Additionally, the table provides detailed information on how the data is split for these two benchmarks \cite{R2GenCMN}.

\begin{table}
\begin{tabular}{l|lll|lll}
\hline
\multirow{2}{*}{Dataset} & \multicolumn{3}{c|}{IU-Xray~\cite{IU-xray}} & \multicolumn{3}{c}{MIMIC-CXR~\cite{mimic_cxr}} \\
\cline{2-7}              & \multicolumn{1}{c}{Train}   & \multicolumn{1}{c}{Val.}       & \multicolumn{1}{c|}{Test}  & \multicolumn{1}{c}{Train}    & \multicolumn{1}{c}{Val.}       & \multicolumn{1}{c}{Test}  \\
\hline
Image                    & 5.2K    & 0.7K       & 1.5K  & 369.0K   & 3.0K       & 5.2K  \\
Report                   & 2.8K    & 0.4K       & 0.8K  & 222.8K   & 1.8K       & 3.3K  \\
Patient                  & 2.8K    & 0.4K       & 0.8K  & 64.6K    & 0.5K       & 0.3K  \\
Avg. Len.                & 37.6    & 36.8       & 33.6  & 53.0     & 53.1       & 66.4  \\
                        \midrule
\end{tabular}
\caption{The specifications of two benchmark datasets that will be utilized to test the COMG model.}
\label{table:dataset}
\end{table}

\subsection{Visualization Results}
\label{Appendix:example_visualization}

This section demonstrates more visualization results predicted by the COMG model. More details are shown in Fig.~\ref{appendix_fig:visualization_prediction}. We have found that the reports generated by our model are more closely related to the ground truth. This includes key organs such as the heart, lungs, and atrium, as well as corresponding diseases such as pleural effusions, pneumothorax, etc.

{\small
\bibliographystyle{ieee_fullname}
\bibliography{egbib}
}

\end{document}